\documentclass[10pt,twocolumn,letterpaper]{article}

\usepackage{cvpr}
\usepackage{times}
\usepackage{epsfig}
\usepackage{graphicx}
\usepackage{amsmath}
\usepackage{amssymb}

\usepackage{algorithmicx,algorithm}

\usepackage{multirow}
\usepackage{makecell}
\usepackage{microtype}
\usepackage{subfigure}
\usepackage{booktabs}

\usepackage[breaklinks=true,bookmarks=false]{hyperref}

\cvprfinalcopy 

\def\cvprPaperID{****} 

\begin{document}

\title{Feature Transformation Ensemble Model with Batch Spectral Regularization 
for Cross-Domain Few-Shot Classification}

\author{
Bingyu Liu, Zhen Zhao, Zhenpeng Li, Jianan Jiang, Yuhong Guo, Jieping Ye\\	
AI Tech, DiDi ChuXing
}

\maketitle
\thispagestyle{empty}

\begin{abstract}
In this paper, we propose a feature transformation ensemble model with batch spectral regularization 
	for the Cross-domain few-shot learning (CD-FSL) challenge. 
Specifically, 
we proposes to construct an ensemble prediction model by  
performing diverse feature transformations 
after a feature extraction network. 
On each branch prediction network of the model we use 
a batch spectral regularization term to suppress the singular values of the feature matrix
during pre-training to improve the generalization ability of the model. 
The proposed model can then be fine tuned in the target domain to address few-shot classification.
We also further apply label propagation, entropy minimization and data augmentation to 
mitigate the shortage of labeled data in target domains. 
Experiments are conducted on a number of CD-FSL benchmark tasks with four target domains 
and the results demonstrate the superiority of our proposed model. 
\end{abstract}

\section{Introduction}

Many current deep learning methods for visual recognition tasks often rely on large amounts of labeled training data to achieve high performance. Collecting and annotating such large training datasets is expensive and impractical in many cases. In order to speed up the research progress, the cross-domain few-shot learning (CD-FSL) challenge~\cite{guo2019new} has been released. It contains data from the CropDiseases,
EuroSAT,
ISIC2018
and ChestX
datasets. The selected datasets can reflect the actual use cases for deep learning.

Meta-learning is a widely used strategy for few-shot learning. However, recent research~\cite{guo2019new} indicates that traditional “pre-training and fine-tuning” can outperform meta-learning based few-shot learning algorithms when there exists a large domain gap between source base classes and target novel classes. Nevertheless, the capacity of fine-tuning can still be limited when facing the large domain gap. To tackle this problem, in the paper, we propose 
a batch spectral regularization (BSR) mechanism to suppress all the singular values of the feature matrix 
in pre-training so that the pre-trained model can avoid overfitting to the source domain and generalize well to the target domain.
Moreover, we propose a feature transformation ensemble model that builds multiple predictors in projected diverse feature
spaces to facilitate cross-domain adaptation and increase prediction robustness.
To mitigate the shortage of labeled data in the target domain, 
we exploit the unlabeled query set in fine-tuning through entropy minimization. 
We also apply a label propagation (LP) step 
to refine the original classification results, and 
exploit data augmentation techniques to augment both the few-shot and test instances from different angles
to improve prediction performance. 
Experiments are conducted on the CD-FSL benchmark tasks 
and the results demonstrate the superiority of our proposed model over the strong fine-tuning baseline.

\section{Approach}

\begin{figure*}[htbp]
\begin{center}
   \includegraphics[width=0.85\linewidth]{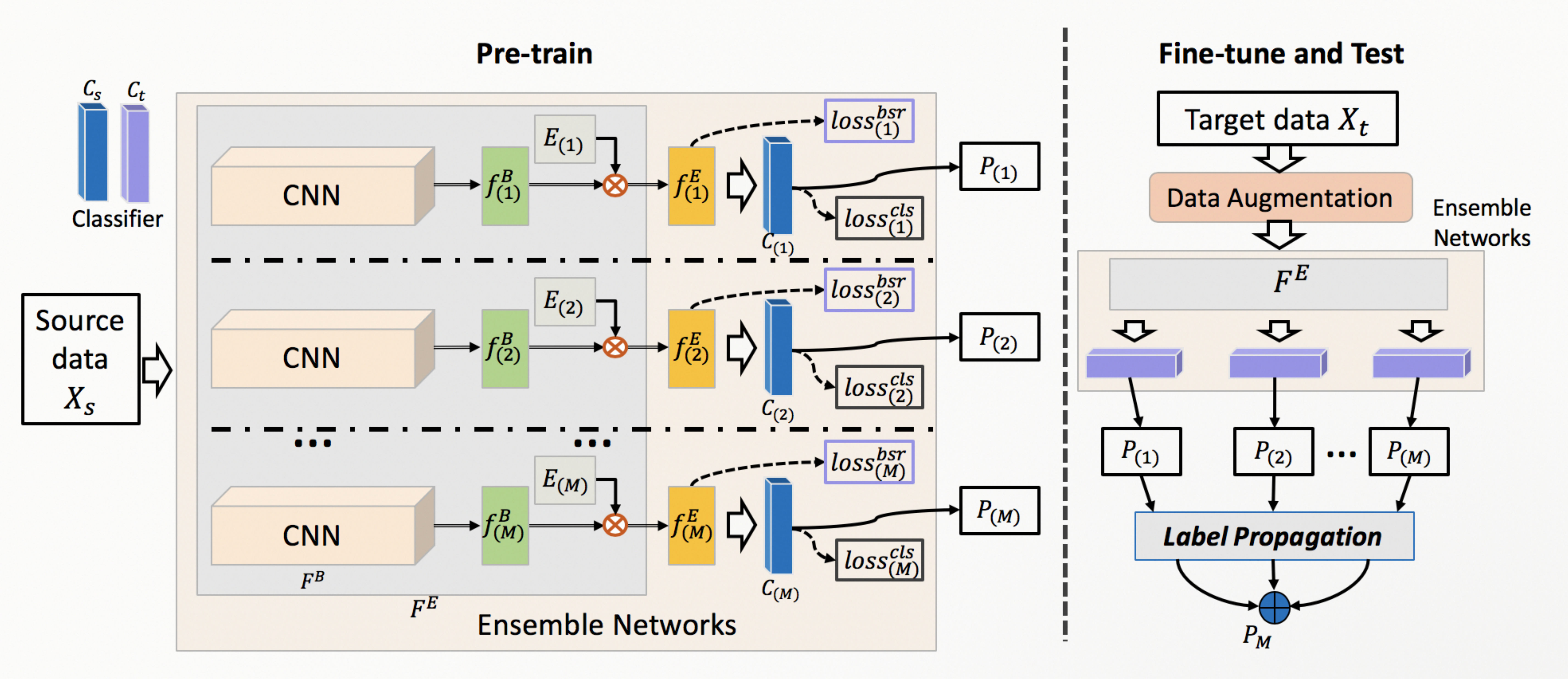}
\end{center}
   \caption{An overview of the proposed approach.}
\label{fig:overview}
\end{figure*}

In the cross-domain few-shot learning setting, we have a source domain $\left( {{X^s},{Y^s}} \right)$ and a target domain $\left( {{X^t},{Y^t}} \right)$. We use all the labeled data in the source domain for pre-training. 
In the target domain, a numbe of 
$K$-way $N$-shot classification tasks are sampled,
each with a support set $S = \left\{ ({x_i},{y_i}) \right\}_{i = 1}^{K \times N}$ composed of $N$ labeled examples from $K$ novel classes.
The labeled support set can be used to fine-tune the pre-trained model, 
while a query set from the same $K$ classes is used to evaluate the model performance. 
Inspired by the ensemble networks for zero-shot learning~\cite{ye2019progressive}, 
we build an ensemble prediction model for cross-domain few-shot learning. 
The overview of our proposed ensemble model is depicted in Fig.~\ref{fig:overview}. 
Below we introduce the components involved in the model.

\subsection{Feature Transformation Ensemble Model}

We build the ensemble model by increasing the diversity of the feature representation space 
while maintaining the usage of the entire training data for each prediction branch network. 
As shown in Fig.~\ref{fig:overview}, 
we use a Convolutional Neural Network (CNN) $F^B$ to extract advanced visual features $f^B\in{\mathbb{R}^{m}}$ from the input data, 
and then transform the features into multiple diverse feature representation spaces using 
different random orthogonal projection matrices 
$\{E_{(1)},E_{(2)}, \cdots ,E_{(M)}\}$ on different branches. 
Each projection matrix $E_{(i)}$ is generated in the follow way. 
We randomly generate a symmetric matrix $Z_{(i)}\in [0,1]^{m\times m}$, 
and then form the orthogonal matrix $E_{(i)}$ using the eigenvectors of $Z_{(i)}$
such that 
$E_{(i)}=[e_1,e_2,...,e_{m-1}]^\top$, 
where $e_i$ represents the eigenvector corresponding to the top $i$-th eigenvalue of $Z_{(i)}$. 
With each projection matrix $E_{(i)}$, we can transform the extracted features into a new feature representation space
such that $f^E_{(i)} = E_{(i)}f^B_{(i)}$, and build a soft-max predictor $C_{(i)}$ in this feature space.  
By using $M$ randomly generated orthogonal projection matrices, we can then build $M$ classifiers. 
In the pre-training stage in the source domain, all the labeled source data is used 
to train each branch network, which includes the composite feature extractor $F^E_{(i)}(x)=E_{(i)}F^B_{(i)}(x)$,
and the classifier $C_{(i)}$, by minimizing the cross-entropy loss. 
In a training batch $(X_B, Y_B)=\{(X_1,Y_1),\cdots,(X_b,Y_b)\}$, 
the loss function can be written as
\begin{equation}
	\ell_{ce}(X_B,Y_B; C_{(i)}\circ F^E_{(i)} ) 
	= \frac{1}{b}\sum\limits_{j = 1}^b L_{ce}\left(C_{(i)}\circ F^E_{(i)}(X_j), Y_j\right) 
\end{equation}
where ${L_{ce}}$ denotes the cross-entropy loss function. 
After pre-training in the source domain, the fine-tuning on the labeled support set of the target domain can be conducted
in the similar way, while the testing on the query instances can be naturally produced in an ensemble manner
by taking the average of the $M$ classifiers' prediction results. 

\subsection{Batch Spectral Regularization}

Previous work~\cite{chen2019catastrophic} shows that penalizing smaller singular values of a feature matrix
can help mitigate negative transfer in fine-tuning.
We extend this penalizer into the full spectrum and 
propose a batch spectral regularization (BSR) mechanism to suppress all the singular values 
of the batch feature matrices
in pre-training,
aiming to avoid overfitting to the source domain and increase generalization ability to the target domain.
This regularization is applied for each branch network of the ensemble model separately in the same way.
For simplicity, we omitted the branch network index in the following presentation. 

Specifically, for a stochastic gradient descent based training algorithm, we work with training batches. 
Given a batch of training instances $(X_B,Y_B)$, its feature matrix can be obtained as 
$A=\left[f^E_1,\cdots, f^E_b\right]$,
where $b$ is the batch size 
and ${f^E_i} = F^E(X_i)$ is the feature vector for the $i$-th instance in the batch.
The BSR term can then be written as 
\begin{equation}
{\ell_{bsr}}\left(A \right) = \sum\limits_{i = 1}^b {\sigma _i^2}
\end{equation}
where ${\sigma _1},{\sigma _2}, \cdots ,{\sigma _b}$ are singular values of the batch feature matrix $A$.
%
The spectral regularized training loss for each batch will be:
\begin{equation}
	\mathcal{L} = \ell_{ce}\left(X_B, Y_B; C\circ F^E\right) 
	+\lambda \ell_{bsr}(F^E(X_B)) 
\end{equation}

\subsection{Label Propagation}

Due to the lack of labeled data in the target domain, the model fine-tuned with the support set can 
can easily make wrong predictions on the query instances. 
Following the effective label refinement procedure in~\cite{ye2017labelless}, 
we propose to apply a label propagation (LP) step 
to exploit the semantic information of unlabeled test data in the extracted feature space
and refine the original classification results. 

Given the prediction score matrix $\hat{Y}^0$ on the query instances $X_Q$ 
with the fine-tuned classifier $C_t$,
we keep the top-$\delta$ fraction of scores in each class (the columns of $\hat{Y}^0$) 
and set other values to zeros in order to propagate only the most confident predictions.
We then build a k-NN graph over the query instances
based on the extracted features $F^E(X_Q)$. 
We use the squared Euclidean distance between each pair of images 
such as $d(i,j) = \left\|F^E(x_i)- F^E(x_j) \right\|^2$ to determine the k-NN graph.
The RBF kernel based affinity matrix $W$ can be computed as follows:
\begin{equation}
{W_{ij}} = \left\{ {\begin{array}{*{20}{c}}
	{\exp \left( \frac{ - d(i,j)}{2{\gamma ^2}} \right),}&{i \in {\rm{KNN}}\left( j \right){\rm{ or\ }}j \in {\rm{KNN}}\left( i \right)}\\
{0,}&{{\rm{otherwise}}}
\end{array}} \right.
\end{equation}
where ${\gamma ^2}$ is the radius of the RBF kernel and ${\rm{KNN}}\left( i \right)$ denotes the k-nearest neighbors of the $i$-th image. 
The normalized Laplacian matrix $L$ can then be calculated as 
$L = {Q^{ - 1/2}}W{Q^{ - 1/2}}$, where $Q$ is a diagonal matrix with ${Q_{ii}} = \sum\nolimits_j {{W_{ij}}} $. 
The label propagation is then performed to provide the following refined prediction score matrix:
\begin{equation}
{Y^*} = {\left( {I - \alpha L} \right)^{ - 1}} \times \hat{Y}^0
\end{equation}
where $I$ is an identity matrix
and $\alpha  \in \left[ {0,1} \right]$ is a trade-off parameter. 
After LP, ${\hat y_i} = \arg {\max _j}Y_{ij}^*$ is used as the predicted class for the $i$-th image.

\subsection{Entropy Minimization}
We extend the semi-supervised learning mechanism into the fine-tuning phase in the target domain
by minimizing the prediction entropy on the unlabeled query set: 
\begin{equation}
{\ell _{ent}}(X_B^q;C \circ {F^E}) =  - \frac{1}{b}\sum\limits_{i = 1}^b {C \circ {F^E}(X_i^q)\log (C \circ {F^E}(X_i^q))}
\end{equation}
where $X_B^q$ denotes a query batch. We can add this term to the original cross-entropy loss 
on the support set batch $(X_B^s,Y_B^s)$ and form a transductive fine-tuning loss function: 
\begin{equation}
{{\cal L}_{ft}} = {\ell _{ce}}(X_B^s,Y_B^s;C \circ {F^E}) + \beta {\ell _{ent}}(X_B^q;C \circ {F^E})
\end{equation}
where $\beta$ is a trade-off parameter.

\subsection{Data Augmentation}

We also exploit data augmentation (DA) strategy with several random operations to supplement 
the support set and make the models learn with more variations. 
In particular, we use combinations of some operations such as image scaling, 
random crop, random flip, random rotation and color jitter to generate a few variants for each image.  
The fine-tuning can be conducted on the augmented support set.
The same augmentation can be conducted for the query set as well,
where several variants of each image can be generated to share the same label. 
Thus the prediction result on each image can be determined by averaging the prediction results
on all the augmented variants of the same image. 

\section{Experiments}

\begin{table}[t]
\caption{Hyper-parameters of augmentation operations.}
\begin{center}
\resizebox{0.85\columnwidth}{!}{
\begin{tabular}{l|c}
\toprule
Augmentation & Hyper-parameters\\ 
\hline
	Scale (S) & $84 \times 84$\\
\hline
	RandomResizedCrop (C) & $84 \times 84$\\
\hline
	\multirowcell{3}{ImageJitter (J)}  & Brightness: 0.4\\
& Contrast:0.4\\
& Color: 0.4\\
\hline
	RandomHorizontalFlip (H) & Flip probability: 50\%\\
\hline
	RandomRotation (R) & $0 - 45$ degrees\\
\bottomrule
\end{tabular}}
\end{center}
\label{table:aug}
\vskip -.1in
\end{table}

\begin{table}[t]
\caption{Compound modes of augmentation operations.}
\begin{center}
\resizebox{0.8\columnwidth}{!}{
\begin{tabular}{l|c}
\toprule
Dataset & Augmentation\\ 
\hline
ISIC \& EuroSAT & \multirowcell{2}{S + SJHR + SR + SJ + SH}\\
\& CropDiseases & \\
\hline
ChestX & S + SJH + C + CJ + CH\\
\bottomrule
\end{tabular}}
\end{center}
\label{table:mode}
\vskip -.1in
\end{table}

\begin{table*}[htbp]
\caption{Results on the CD-FSL benchmark.}
\begin{center}
\resizebox{\textwidth}{!}{
\begin{tabular}{lccc|ccc}
\toprule
\multirowcell{2}{Methods} & \multicolumn{3}{c|}{ChestX} & \multicolumn{3}{c}{ISIC} \\ 
\cline{2-7} 
& 5-way 5-shot & 5-way 20-shot & 5-way 50-shot & 5-way 5-shot & 5-way 20-shot & 5-way 50-shot \\ 
\cline{1-7}
Fine-tuning~\cite{guo2019new} & 25.97\%\(\pm\)0.41\% & 31.32\%\(\pm\)0.45\% & 35.49\%\(\pm\)0.45\% & 48.11\%\(\pm\)0.64\% & 59.31\%\(\pm\)0.48\% & 66.48\%\(\pm\)0.56\%\\ \hline
BSR & 26.84\%\(\pm\)0.44\% & 35.63\%\(\pm\)0.54\% & 40.18\%\(\pm\)0.56\% & 54.42\%\(\pm\)0.66\% & 66.61\%\(\pm\)0.61\% & 71.10\%\(\pm\)0.60\%\\
BSR+LP & 27.10\%\(\pm\)0.45\% & 35.92\%\(\pm\)0.55\% & 40.56\%\(\pm\)0.56\% & 55.86\%\(\pm\)0.66\% & 67.48\%\(\pm\)0.60\% & 72.17\%\(\pm\)0.58\%\\
BSR+DA & 28.20\%\(\pm\)0.46\% & 36.72\%\(\pm\)0.51\% & 42.08\%\(\pm\)0.53\% & 54.97\%\(\pm\)0.68\% & 66.43\%\(\pm\)0.57\% & 71.62\%\(\pm\)0.60\%\\
	BSR+LP+ENT & 26.86\%\(\pm\)0.45\% & 35.60\%\(\pm\)0.51\% & 42.26\%\(\pm\)0.53\% & {\bf 56.82}\%\(\pm\)0.68\% & {\bf 68.97}\%\(\pm\)0.56\% & {\bf 74.13}\%\(\pm\)0.56\%\\
	BSR+LP+DA & {\bf 28.50}\%\(\pm\)0.48\% & {\bf 36.95}\%\(\pm\)0.52\% & {\bf 42.32}\%\(\pm\)0.53\% & 56.25\%\(\pm\)0.69\% & 67.31\%\(\pm\)0.57\% & 72.33\%\(\pm\)0.58\%\\
\cline{1-7}
BSR (Ensemble) & 28.44\%\(\pm\)0.45\% & 37.05\%\(\pm\)0.50\% & 43.22\%\(\pm\)0.54\% & 55.47\%\(\pm\)0.68\% & 68.00\%\(\pm\)0.59\% & 73.36\%\(\pm\)0.54\%\\
BSR+LP (Ensemble) & 28.66\%\(\pm\)0.44\% & 37.44\%\(\pm\)0.51\% & 43.72\%\(\pm\)0.54\% & 57.14\%\(\pm\)0.67\% & 68.99\%\(\pm\)0.58\% & 74.62\%\(\pm\)0.54\%\\
BSR+DA (Ensemble) & 29.09\%\(\pm\)0.45\% &  37.89\%\(\pm\)0.53\% & 43.98\%\(\pm\)0.56\%  & 56.13\%\(\pm\)0.66\% & 67.10\%\(\pm\)0.61\% &   73.16\%\(\pm\)0.54\%\\
	BSR+LP+ENT (Ensemble) & 28.00\%\(\pm\)0.46\% & 36.87\%\(\pm\)0.52\% & 43.79\%\(\pm\)0.53\% & {\bf 58.02}\%\(\pm\)0.68\% & {\bf 70.22}\%\(\pm\)0.59\% & {\bf 74.94}\%\(\pm\)0.55\%\\
	BSR+LP+DA (Ensemble) & {\bf 29.72}\%\(\pm\)0.45\% &  {\bf 38.34}\%\(\pm\)0.53\% & {\bf 44.43}\%\(\pm\)0.56\%  & 57.40\%\(\pm\)0.67\% & 68.09\%\(\pm\)0.60\% &   74.08\%\(\pm\)0.55\%\\
\bottomrule
\end{tabular}}
\resizebox{\textwidth}{!}{
\begin{tabular}{lccc|ccc}
\toprule
\multirowcell{2}{Methods} & \multicolumn{3}{c|}{EuroSAT} & \multicolumn{3}{c}{CropDiseases} \\ 
\cline{2-7} 
& 5-way 5-shot & 5-way 20-shot & 5-way 50-shot & 5-way 5-shot & 5-way 20-shot & 5-way 50-shot \\ 
\cline{1-7}
Fine-tuning~\cite{guo2019new} & 79.08\%\(\pm\)0.61\% & 87.64\%\(\pm\)0.47\% & 90.89\%\(\pm\)0.36\% & 89.25\%\(\pm\)0.51\% & 95.51\%\(\pm\)0.31\% & 97.68\%\(\pm\)0.21\%\\ \hline
BSR & 80.89\%\(\pm\)0.61\% & 90.44\%\(\pm\)0.40\% & 93.88\%\(\pm\)0.31\% & 92.17\%\(\pm\)0.45\% & 97.90\%\(\pm\)0.22\% & 99.05\%\(\pm\)0.14\%\\
BSR+LP & 84.35\%\(\pm\)0.59\% & 91.99\%\(\pm\)0.37\% & 95.02\%\(\pm\)0.27\% & 94.45\%\(\pm\)0.40\% & 98.65\%\(\pm\)0.19\% & 99.38\%\(\pm\)0.11\%\\
BSR+DA & 82.75\%\(\pm\)0.55\% & 92.61\%\(\pm\)0.31\% & 95.26\%\(\pm\)0.39\% & 93.99\%\(\pm\)0.39\% & 98.62\%\(\pm\)0.15\% & 99.42\%\(\pm\)0.08\%\\
BSR+LP+ENT & 85.70\%\(\pm\)0.53\% & 92.90\%\(\pm\)0.33\% & 95.40\%\(\pm\)0.29\% & 95.69\%\(\pm\)0.35\% & 98.60\%\(\pm\)0.18\% & 99.27\%\(\pm\)0.12\%\\
	BSR+LP+DA & {\bf 85.97}\%\(\pm\)0.52\% & {\bf 93.73}\%\(\pm\)0.29\% & {\bf 96.07}\%\(\pm\)0.30\% & {\bf 95.97}\%\(\pm\)0.33\% & {\bf 99.10}\%\(\pm\)0.12\% & {\bf 99.66}\%\(\pm\)0.07\%\\
\cline{1-7}
BSR (Ensemble) & 83.93\%\(\pm\)0.53\% & 92.55\%\(\pm\)0.33\% & 95.11\%\(\pm\)0.24\% & 93.54\%\(\pm\)0.41\% & 98.34\%\(\pm\)0.20\% & 99.22\%\(\pm\)0.12\%\\
BSR+LP (Ensemble) & 86.08\%\(\pm\)0.55\% & 93.81\%\(\pm\)0.30\% & 95.97\%\(\pm\)0.23\% & 95.48\%\(\pm\)0.38\% & 98.94\%\(\pm\)0.16\% & 99.49\%\(\pm\)0.11\%\\
BSR+DA (Ensemble) & 85.19\%\(\pm\)0.51\% & 93.68\%\(\pm\)0.28\% &  96.14\%\(\pm\)0.26\% & 94.80\%\(\pm\)0.36\% & 98.69\%\(\pm\)0.16\% &    99.51\%\(\pm\)0.09\%\\
BSR+LP+ENT (Ensemble) & 87.17\%\(\pm\)0.52\% & 93.96\%\(\pm\)0.29\% & 96.09\%\(\pm\)0.22\% & 96.04\%\(\pm\)0.36\% & 98.94\%\(\pm\)0.16\% & 99.45\%\(\pm\)0.10\%\\
	BSR+LP+DA (Ensemble) & {\bf 88.13}\%\(\pm\)0.49\% & {\bf 94.72}\%\(\pm\)0.28\% &  {\bf 96.89}\%\(\pm\)0.19\% & {\bf 96.59}\%\(\pm\)0.31\% & {\bf 99.16}\%\(\pm\)0.14\% &  {\bf 99.73}\%\(\pm\)0.06\%\\
\bottomrule
\end{tabular}}
\end{center}
\label{table:result}
\vskip -.1in
\end{table*}

\begin{table}[htbp]
\caption{Averages across all datasets and shot levels.}
\begin{center}
\resizebox{0.7\columnwidth}{!}{
\begin{tabular}{l|c}
\toprule
Methods & Average\\ 
\hline
Fine-tuning~\cite{guo2019new} & 67.23\%\(\pm\)0.46\%\\ \hline
BSR & 70.76\%\(\pm\)0.46\%\\
BSR+LP & 71.91\%\(\pm\)0.44\%\\
BSR+DA & 71.89\%\(\pm\)0.44\%\\
BSR+LP+ENT & 72.68\%\(\pm\)0.42\%\\
BSR+LP+DA & 72.85\%\(\pm\)0.42\%\\
\hline
BSR (Ensemble) & 72.35\%\(\pm\)0.43\%\\
BSR+LP (Ensemble) & 73.36\%\(\pm\)0.42\%\\
BSR+DA (Ensemble) & 72.95\%\(\pm\)0.42\%\\
BSR+LP+ENT (Ensemble) & 73.62\%\(\pm\)0.42\%\\
BSR+LP+DA (Ensemble) & 73.94\%\(\pm\)0.40\%\\
\bottomrule
\end{tabular}}
\end{center}
\label{table:avg}
\vskip -.2in
\end{table}

\subsection{Setup}

In the experiments, we use the evaluation protocol in the CD-FSL challenge~\cite{guo2019new}, which 
takes 15 images from each class as the query set
and uses 600 randomly sampled few-shot episodes in each target domain, 
The average accuracy and 95\% confidence interval are reported.

As for the model architecture, we use ResNet-10
as the CNN feature extractor ${F^B}$ and a fully-connected layer with soft-max activation
as the classifier ${C}$. 
We set the trade-off parameters $\lambda  = 0.001$, $\beta=0.1$ and the number of branches $M = 10$. 
For the label propagation step, we use $k= 10$ for the k-NN graph construction, and set ${\gamma ^2}$ as the average of the squared distances of the edges in the k-NN graph. The parameters $\delta$ and $\alpha$ are set to 0.2 and 0.5 respectively.
We adopt mini-batch SGD with momentum of 0.9 for both pre-training and fine-tuning. During the pre-training stage, models are trained for 400 epochs. The learning rate and the weight decay are set to 0.001 and 0.0005 respectively. During fine-tuning, we set the learning rate to 0.01 and fine-tune for 100 epochs.

For data augmentation (DA), we choose 5 types of augmentation operations 
as shown in Table~\ref{table:aug}. 
We use 5 compound modes of these operations to generate data
in different target domains. The specific operations used in each target domain are shown in Table~\ref{table:mode}.

\subsection{Results}
We investigate a number of variants of the proposed model by comparing with the strong fine-tuning baseline result reported 
in~\cite{guo2019new}. 
We first investigate a single prediction network with batch spectral regularization (BSR) without ensemble
and its other variants that further incorporate label propagation (LP) or/and data augmentation (DA). 
Then we extend these variants into the ensemble model framework with $M=10$. 
The results are reported in Table~\ref{table:result}, and
the average accuracies (and 95\% confidence internals) across all datasets and shot levels are shown 
in Table~\ref{table:avg}. 

We can see that even with only BSR, the proposed method can already significantly outperform the fine-tuning baseline 
(average 70.76\% vs 67.23\%).  
The ensemble BSR further improves the results (72.35\%).
The LP and DA components can also help improve the CD-FSL performance.
We observe that on target domains more similar to the source domain, LP performs better than DA and vice versa. 
This shows that LP and DA focus on different aspects of the data.
As a result combining LP and DA can further improve the performances. 
Moreover, DA is not very effective for the ISIC domain, where it even degrades the performance in some cases.
Also the experiment's running time is typically longer with DA. 
By replacing DA with ENT, we can obtain similar overall performance.
With a single model, the best average result achieved by BSR+LP+DA is 72.85\%,
while BSR+LP+ENT achieves 72.68\%.
With the ensemble model, BSR+LP+DA (Ensemble) produces the best average result 73.94\%,
while BSR+LP+ENT (Ensemble) yields 73.62\%.

\section{Conclusion}
In this paper, we proposed a feature transformation based ensemble model for CD-FSL. 
The model also incorporates batch spectral regularization in pre-training, and exploits data augmentation 
and label propagation during fine-tuning and testing in the target domain. 
The combinational models produced superior CD-FSL results comparing to the strong fine-tuning baseline.

{\small
\bibliographystyle{ieee_fullname}
\bibliography{egbib}

\begin{thebibliography}{1}\itemsep=-1pt

\bibitem{chen2019catastrophic}
Xinyang Chen, Sinan Wang, Bo Fu, Mingsheng Long, and Jianmin Wang.
\newblock Catastrophic forgetting meets negative transfer: Batch spectral
  shrinkage for safe transfer learning.
\newblock In {\em NeurIPS}, 2019.

\bibitem{guo2019new}
Yunhui Guo, Noel~CF Codella, Leonid Karlinsky, John~R Smith, Tajana Rosing, and
  Rogerio Feris.
\newblock A new benchmark for evaluation of cross-domain few-shot learning.
\newblock {\em arXiv preprint arXiv:1912.07200}, 2019.

\bibitem{ye2017labelless}
Meng Ye and Yuhong Guo.
\newblock Labelless scene classification with semantic matching.
\newblock In {\em BMVC}, 2017.

\bibitem{ye2019progressive}
Meng Ye and Yuhong Guo.
\newblock Progressive ensemble networks for zero-shot recognition.
\newblock In {\em CVPR}, 2019.

\end{thebibliography}
}

\end{document}